\documentclass[runningheads]{llncs}

\newcommand {\ent} {\mathrel{{\scriptstyle\mid\!\sim}}}

\newcommand {\sx} {\langle}
\newcommand {\dx} {\rangle}

\newcommand {\emme} {\mathcal{M}}

\newcommand {\tc} {\mid}
\newcommand {\vuoto} {\emptyset}

\newcommand{\tip}{{\bf T}}

\newcommand{\el}{\mathcal{EL}^{\bot}}
\newcommand{\elpb}{{\mathcal{EL}}^{+}_{\bot}}

\newcommand{\eltm}{\mathcal{EL}^{\bot} \tip_{min}}

\newcommand{\dlltm}{\mbox{\em DL-Lite}_{\mathit{c}}\tip_{min}}

\newcommand{\be}{\begin{enumerate}}
\newcommand{\ee}{\end{enumerate}}

\newcommand{\hide}[1]{}

\def \cases{\left \{\begin{array}{l}}
\def \endcases{\end{array}\right .}

\newcommand {\bes} {\begin{description}}
\newcommand{\ens} {\end{description}}

\newcommand {\beq} {\begin{quote}}
\newcommand {\enq} {\end{quote}}
\newcommand {\bit} {\begin{itemize}}
\newcommand {\enit} {\end{itemize}}

\newenvironment{pozz}{\color{black}}{\color{black}}
\usepackage{times}
\usepackage{helvet}
\usepackage{courier}
\usepackage{times}
\usepackage{graphicx}
\usepackage{latexsym}
\usepackage{graphicx}
\usepackage{color}
\usepackage{amssymb}
\usepackage{colortbl}
\usepackage{amsmath}
\usepackage{microtype}
\usepackage{amsmath}

\begin{document}
\bibliographystyle{plain}

\title{
On a plausible concept-wise multipreference semantics \\
and its relations with self-organising maps}



\author{Laura Giordano \inst{1} \and Valentina Gliozzi \inst{2} \and Daniele Theseider Dupr{\'{e}}  \inst{1}}

\institute{DISIT - Universit\`a del Piemonte Orientale, 
 Alessandria, Italy \and
Dipartimento di Informatica,
Universit\`a di Torino, Italy, 
}

\authorrunning{ }
\titlerunning{ }

 \maketitle
 

\begin{abstract}In this paper we describe a concept-wise multi-preference semantics for description logic which has its root in the preferential approach for modeling defeasible reasoning in knowledge representation. We argue that this proposal, beside satisfying some desired properties, such as KLM postulates, and avoiding the drowning problem, also defines a plausible notion of semantics.
We motivate the plausibility of the concept-wise multi-preference semantics by developing a logical semantics of  self-organising maps, which have been proposed as possible candidates to explain the psychological mechanisms underlying category generalisation, in terms of multi-preference interpretations.



\end{abstract}


\vspace{-0.1cm}
\section{Introduction}
\vspace{-0.1cm}

Conditional logics have 
have their roots in philosophical logic.
They have  been studied
first by Lewis \cite{Lewis:73,Nute80}  to formalize 
hypothetical and counterfactual reasoning (if $A$ were the case then $B$) that
cannot be captured by classical  logic with its material
implication. From the 80's they have been considered in
computer science and artificial intelligence 
and they have provided an axiomatic foundation of non-monotonic  and 
common sense reasoning \cite{Delgrande:87,KrausLehmannMagidor:90}.
%
In particular, preferential approaches  \cite{KrausLehmannMagidor:90,whatdoes} to common sense reasoning
have been more recently extended to description logics, to deal with inheritance with exceptions in ontologies,
allowing for non-strict forms of inclusions,
called {\em typicality or defeasible inclusions} (namely, conditionals), with different preferential semantics \cite{lpar2007,sudafricaniKR} 
and closure constructions \cite{casinistraccia2010,CasiniDL2013,dl2013,Pensel18}.

In this paper we consider a ``concept-aware" multipreference semantics \cite{arXiv_iclp2020} that has been recently introduced for a lightweight description logic of the $\el$ family, which takes into account preferences with respect to different concepts, and integrates them into a preferential semantics. 
To support the plausibility of this semantics we show that 
it can be can used to provide a logical semantics of self-organising maps \cite{kohonen2001}.
Self-organising maps (SOMs) have been proposed as possible candidates to explain the psychological mechanisms underlying category generalisation. They are psychologically and biologically plausible neural network models that can learn after limited exposure to positive category examples, without any need of contrastive information.

We show that the process of category generalization in self-organising maps produces, as a result, a multipreference model
in which a preference relation is associated to each concept (each learned category) and  the combination of the preferences into a global one, following the approach in \cite{arXiv_iclp2020}, defines a standard KLM preferential model.
The model can be used to learn or validate conditional knowledge from the empirical data used in the category generalization process,
and the evaluation of conditionals can be done by model checking, using the information recorded in the SOM.

Based on the assumption that the abstraction process in the SOM is able to identify the most typical exemplars for a given category,
in the semantic representation of a category, we will identify some specific exemplars (namely, the best matching units 
of the category) as the typical exemplars of the category, thus defining a preference relation among the instances of a category.  


The category generalization process can then be regarded as a model building process and, in a way, as a belief revision process.
Indeed, initially we have no belief about which is the category of any exemplar.
During training, the current state of the SOM corresponds to a model representing the beliefs about the input exemplars considered so far (concerning their category).
Each time a new input exemplar is considered, this model is revised adding the exemplar into the proper category.

\section{Preliminary: the description logic $\el$}  \label{sec:EL}

We consider the description logic $\el$ of the ${\cal EL}$ family  \cite{rifel}. 
Let ${N_C}$ be a set of concept names, ${N_R}$ a set of role names
  and ${N_I}$ a set of individual names.  
The set  of $\el$ \emph{concepts} can be
defined as follows: 
  $C \ \ := A \tc \top \tc \bot  \tc C \sqcap C \tc \exists r.C $, 
where $a \in N_I$, $A \in N_C$ and $r \in N_R$. 
Observe that union, complement and universal restriction are not $\el$ constructs.
A knowledge base (KB) $K$ is a pair $({\cal T}, {\cal A})$, where ${\cal T}$ is a TBox and
${\cal A}$ is an ABox.
The TBox ${\cal T}$ is  a set of {\em concept inclusions} (or subsumptions) of the form $C \sqsubseteq D$, where $C,D$ are concepts.
The  ABox ${\cal A}$ is  a set of assertions of the form $C(a)$ 
and $r(a,b)$ where $C$ is a  concept, $r \in N_R$, and $a, b \in N_I$.

An {\em interpretation} for $\el$ is a pair $I=\langle \Delta, \cdot^I \rangle$ where:
$\Delta$ is a non-empty domain---a set whose elements are denoted by $x, y, z, \dots$---and 
$\cdot^I$ is an extension function that maps each
concept name $C\in N_C$ to a set $C^I \subseteq  \Delta$, each role name $r \in N_R$
to  a binary relation $r^I \subseteq  \Delta \times  \Delta$,
and each individual name $a\in N_I$ to an element $a^I \in  \Delta$.
It is extended to complex concepts  as follows:
$\top^I=\Delta$, $\bot^I=\vuoto$, 
$(C \sqcap D)^I =C^I \cap D^I$  and 
$(\exists r.C)^I =\{x \in \Delta \tc \exists y.(x,y) \in r^I \ \mbox{and} \ y \in C^I\}.$	

\noindent
The notions of satisfiability of a KB  in an interpretation and of entailment are defined as usual:

\begin{definition}[Satisfiability and entailment] \label{satisfiability}
Given an $\el$ interpretation $I=\langle \Delta, \cdot^I \rangle$: 

	- $I$  satisfies an inclusion $C \sqsubseteq D$ if   $C^I \subseteq D^I$;

	-  $I$ satisfies an assertion $C(a)$ if $a^I \in C^I$ and an assertion $r(a,b)$ if $(a^I,b^I) \in r^I$.

\noindent
 Given  a KB $K=({\cal T}, {\cal A})$,
 an interpretation $I$  satisfies ${\cal T}$ (resp. ${\cal A}$) if $I$ satisfies all  inclusions in ${\cal T}$ (resp. all assertions in ${\cal A}$);
 $I$ is a \emph{model} of $K$ if $I$ satisfies ${\cal T}$ and ${\cal A}$.

 A subsumption $F= C \sqsubseteq D$ (resp., an assertion $C(a)$, $R(a,b)$),   {is entailed by $K$}, written $K \models F$, if for all models $I=$$\sx \Delta,  \cdot^I\dx$ of $K$,
$I$ satisfies $F$.
\end{definition}

\section{A concept-wise multi-preference semantics} \label{sec:multipref}

In this section  we describe an extension of $\el$ with typicality inclusions, defined along the lines of the extension of description logics with typicality \cite{lpar2007,AIJ15}, but we exploit a different multi-preference semantics \cite{arXiv_iclp2020}.
In addition to standard $\el$ inclusions $C \sqsubseteq D$ (called  {\em strict} inclusions in the following), the TBox ${\cal T}$ will also contain typicality inclusions of the form $\tip(C) \sqsubseteq D$, where $C$ and $D$ are $\el$ concepts. 
A typicality inclusion $\tip(C) \sqsubseteq D$ means that ``typical C's are D's" or ``normally C's are D's" and corresponds to a conditional implication $C \ent D$ in Kraus, Lehmann and Magidor's (KLM) preferential approach \cite{KrausLehmannMagidor:90,whatdoes}. 
Such inclusions are defeasible, i.e.,  admit exceptions, while 
strict inclusions must be satisfied by all domain elements.

Let ${\cal C}= \{C_1, \ldots, C_k\}$ be a set of distinguished $\el$ concepts. 
For each concept $C_i \in {\cal C}$, we introduce a modular preference relation $<_{C_i}$ which describes the preference among domain elements with respect to $C_i$.
Each preference relation $<_{C_i}$ has the same properties of preference relations in KLM-style ranked interpretations \cite{whatdoes}, is a modular and well-founded partial order, i.e., irreflexive and transitive relation, where: $<_{C_i}$ is {\em well-founded} 
if, for all $S \subseteq \Delta$, if $S\neq \emptyset$, then $min_{<_{C_i}}(S)\neq \emptyset$;
and  $<_{C_i}$ is {\em modular} if,
for all $x,y,z \in \Delta$, if $x <_{C_j} y$ then $x <_{C_j} z$ or $z <_{C_j} y$).

  \begin{definition}[Multipreference interpretation]\label{interpretazione_Ci}  
A {\em multipreference interpretation}  is a tuple
$\emme_{C_i}= \langle \Delta, <_{C_1}, \ldots, <_{C_k}, \cdot^I \rangle$, 
where:
\begin{itemize}
\item[(a)] $\Delta$ is a non-empty domain;
 
\item[(b)] $<_{C_i}$ is an irreflexive, transitive, well-founded and modular relation over $\Delta$;

\item[(d)]  
$\cdot^I$ is an interpretation function, as in an $\el$ interpretation  
(see Section \ref{sec:EL}).
\end{itemize}

\end{definition}
Observe that, given a multipreference interpretation, an interpretation $\emme_{C_i}= \langle \Delta, <_{C_i}, \cdot^I \rangle$ can be associated to each concept $C_i$, which is a ranked interpretation as those considered for $\el$ plus typicality in \cite{TPLP2016}. The preference relation $<_{C_i}$ allows the set of prototypical  $C_i$-elements to be defined as the $C_i$-elements which are minimal with respect to $<_{C_i}$, i.e., $min_{<_{C_i}} (C_i^I)$.
As a consequence, the multipreference interpretation above is able to single out the typical $C_i$-elements, for all distinguished concepts $C_i \in {\cal C}$.

The multipreference  structures above are at the basis of the semantics for ranked $\el$ knowledge bases  \cite{arXiv_iclp2020}, which have been
inspired to Brewka's framework of basic preference descriptions  \cite{Brewka04}. 
A {\em ranked TBox}  ${\cal T}_{C_i}$ is allowed for each concept $C_i \in {\cal C}$,  and contains all the defeasible inclusions, $\tip(C_i) \sqsubseteq D$, specifying the typical properties of $C_i$-elements. 
Ranks (non-negative integers) are assigned to such inclusions; the ones with higher ranks are considered to be more important than the ones with lower ranks.

Consider, for instance, the ranked knowledge base $K =\langle {\cal T}_{strict},  {\cal T}_{Employee}, {\cal T}_{Student},$ ${\cal T}_{PhDStudent}, {\cal A} \rangle$, over the set of distinguished concepts ${\cal C}=\{\mathit{Employee, Student,}$ $\mathit{PhDStudent}\}$, with empty ABox,
and with $ {\cal T}_{strict}$ the set of strict inclusions:

$\mathit{Employee  \sqsubseteq  Adult}$ \ \ \ \ \ \ \ \ \ \ \ \ \ \ \   $\mathit{Adult  \sqsubseteq  \exists has\_SSN. \top}$  
 \ \ \ \ \ \ \ \ \ \ \   $\mathit{PhdStudent  \sqsubseteq  Student}$ 

$\mathit{Young  \sqcap NotYoung \sqsubseteq  \bot}$ \ \ \ \ \ \ \ \
$\mathit{\exists hasScholarship.\top  \sqcap Has\_no\_Scholarship \sqsubseteq  \bot}$;

\noindent
the ranked TBox ${\cal T}_{Employee} =\{(d_1,0), (d_2,0)\}$ contains the defeasible inclusions:

$(d_1)$ $\mathit{\tip(Employee) \sqsubseteq NotYoung}$  \ \ \ \ \ \ \ \ \  \ \ \ \ \  

$(d_2)$ $\mathit{\tip(Employee) \sqsubseteq \exists has\_boss.Employee}$;

\noindent
the ranked TBox ${\cal T}_{Student}= \{(d_3,0),(d_4,1), (d_5,1)\}$ contains the defeasible inclusions:

$(d_3)$ $\mathit{\tip(Student) \sqsubseteq  \exists has\_classes.\top}$   \ \ \ \ \ \ \ \ \ \  

$(d_4)$ $\mathit{\tip(Student) \sqsubseteq Young}$

$(d_5)$ $\mathit{\tip(Student) \sqsubseteq  Has\_no\_Scholarship}$

\noindent
and the ranked TBox ${\cal T}_{PhDStudent}=\{  (d_6, 0), (d_7,1)\}$ contains the inclusions:

$(d_6)$ $\mathit{\tip(PhDStudent) \sqsubseteq  \exists hasScholarship.Amount}$ \ \ \ \ \ \ \ \  

$(d_7)$ $\mathit{\tip(PhDStudent) \sqsubseteq Bright}$




%
%
%

\noindent
Exploiting the fact that for an $\el$ knowledge base we can restrict our consideration to finite domains \cite{rifel}, and considering canonical models which are large enough to contain a domain element for each possible consistent concept occurring in $K$ (and its complement),
the ranked knowledge base $K$ above gives rise to canonical models, where the three preference relations $<_\mathit{Employee}$, $<_\mathit{Student}$, and $<_\mathit{PhDStudent}$ represent the preference among the elements of the domain $\Delta$ according to concepts  $\mathit{Employee}$, $\mathit{Student}$, and $\mathit{PhDStudent}$, respectively. 

While we refer to  \cite{arXiv_iclp2020} for the construction of the preference relations $<_{C_i}$'s 
from a ranked knowledge base $K$, in the following we will recall the notion of concept-wise multi-preference interpretation which can be obtained 
by {\em combining} the preference relations $<_{C_i}$ into a global preference relation $<$. 
This is needed for reasoning about the typicality of arbitrary $\el$ concepts $C$, 
which do not belong to the set of distinguished concepts ${\cal C}$.
For instance,  we may want to verify whether typical employed students are young, or whether they have a boss. To answer these questions both preference relations $<_\mathit{Employee}$ and $<_\mathit{Student}$ are relevant, and they might be conflicting for some pairs of domain elements as, for instance, tom is more typical than bob as a student ($\mathit{tom <_\mathit{Student} bob}$), but more exceptional as an employee ( $\mathit{bob <_\mathit{Employee} tom}$).

To define a global preference relation, 
we take into account the specificity relation among concepts,
such as, for instance, the fact that a concept like $\mathit{PhdStudent}$ is more specific than concept $\mathit{Student}$. 
The idea is that,  in case of conflicts, the properties of a more specific class (such as that PhD students normally have a scholarship) should  override the properties of less specific class (such as that students normally do not have a scholarship).
\begin{definition}[Specificity] \label{specificity}
A {\em specificity relation} among concepts in ${\cal C}$ is a binary relation 
$\succ \subseteq {\cal C} \times {\cal C}$ which is irreflexive and transitive.
\end{definition}
For $C_h, C_j \in {\cal C}$, $C_h \succ C_j$ means that $C_h$ is {\em more specific than}  $C_j$.
The simplest notion of {\em specificity} among concepts with respect to a knowledge base $K$ is based on the subsumption hierarchy:  
 $C_h \succ C_j$
  if $ {\cal T}_{strict} \models_{\el} C_h \sqsubseteq C_j$  and $ {\cal T}_{strict} \not\models_{\el} C_j \sqsubseteq C_h$. 
This is one of the notions of specificity considered  for ${\cal DL}^N$  \cite{bonattiAIJ15}. 
Another one is based on the ranking of concepts in the rational closure of $K$.

%
%
%

Let us recall the notion of concept-wise multipreference interpretation \cite{arXiv_iclp2020}. 
 \begin{definition}[concept-wise multipreference interpretation]\label{def-multipreference-int}  
A {\em concept-wise multipreference interpretation} (or cw$^m$-interpretation) is a tuple $\emme= \langle \Delta, <_{C_1}, \ldots,<_{C_k}, <, \cdot^I \rangle$
such that:   
\begin{itemize}
 
\item[(a)]  $\Delta$ is a non-empty domain; 

\item[(b)] for each $i=1,\ldots, k$, $<_{C_i}$ is an irreflexive, transitive, well-founded and modular relation over $\Delta$; 

\item[(c)]  $<$ is a (global) preference relation over $\Delta$ defined from  $<_{C_1}, \ldots,<_{C_k}$ as follows: \ \ 
%
\begin{align*}
x <y  \mbox{ iff \ \ } 
(i) &\  x <_{C_i} y, \mbox{ for some } C_i \in {\cal C}, \mbox{ and } \\
(ii) & \ \mbox{  for all } C_j\in {\cal C}, \;  x \leq_{C_j} y \mbox{ or }  \exists C_h (C_h \succ C_j  \mbox{ and } x <_{C_h} y )
\end{align*}
\item[(d)]  $\cdot^I$ is an interpretation function, as defined for $\el$ interpretations 
(see Section \ref{sec:EL}),
with the addition that, for typicality concepts, we let: 
$$(\tip(C))^I = min_{<}(C^I)$$
where $Min_<(S)= \{u: u \in S$ and $\nexists z \in S$ s.t. $z < u \}$.

\end{itemize}
\end{definition}
Relation $<$ is defined from $<_{C_1}, \ldots,<_{C_k}$  based on a {\em modified} Pareto condition:
$x< y$ holds if there is at least a $C_i \in {\cal C}$ such that $ x <_{C_i} y$ and,
 for all $C_j \in {\cal C}$,   either $x \leq_{C_j} y$ holds or, in case it does not, there is some $C_h$ more specific than $C_j$ such that $x <_{C_h} y$ (preference  $<_{C_h}$ in this case overrides $<_{C_j}$).
 The idea is that, for two PhD students (who are also students) Bob and Mary, if $\mathit{mary <_\mathit{Student} bob}$ and $\mathit{bob <_\mathit{PhDStudent} mary}$, we will have $\mathit{bob < mary}$, that is, Bob is regarded as being globally more typical than Mary 
 as he satisfies more properties of typical PhD students wrt Mary 
 although Mary may satisfy additional properties of typical students wrt Bob. 

It has been proven \cite{arXiv_iclp2020} that, given a cw$^m$-interpretation $\emme= \langle \Delta, <_{C_1}, \ldots,<_{C_k}, <, \cdot^I \rangle$,
the relation $<$ is an irreflexive, transitive and well-founded relation. Hence, the triple $\emme'= \langle \Delta,  <, \cdot^I \rangle$ is a KLM-style preferential interpretation, as those introduced for $\el$ with typicality \cite{ijcai2011} (and it is not necessarily a modular interpretation). A cw$^m$-model of a ranked $\el$ knowledge base $K$ is then defined as a specific preferential interpretation which builds over the preference relations $<_{C_i}$, constructed from the ranked TBoxes ${\cal T}_{C_i}$, and satisfies all strict inclusions and assertions in $K$.
The notion of cw$^m$-entailment, defined in the obvious way, satisfies the KLM postulates of a preferential consequence relation, and does not suffer from the drowning problem.  
In the next section we motivate the plausibility of this concept-wise multipreference semantics showing that it is well suited to provide a semantic characterization of self-organising maps \cite{kohonen2001}.

\section{Self-organising maps}\label{som-general}

%
%

Self-organising maps (SOMs, introduced by Kohonen \cite{kohonen2001}) are particularly plausible neural network models that learn in a human-like manner. In particular: SOMs learn to organize {stimuli into} categories in an {\em unsupervised} way, without the need of a teacher providing a feedback;
can learn with just {a few} positive stimuli, without the need for negative examples or contrastive information; reflect basic constraints of a plausible brain implementation in different areas of the cortex \cite{miikkulainen2005}, and are therefore biologically plausible models of category formation; have proven to be capable of explaining experimental results. 

In this section we shortly describe the architecture of SOMs and report Gliozzi and Plunkett'  similarity-based account of category generalization based on SOMs \cite{CogSci2017}. Roughly speaking, in  \cite{CogSci2017} the authors judge a new stimulus as belonging  to a category by comparing the distance of the stimulus from the category representation to the precision of the category representation. 

SOMs  consist of a set of neurons, or units, spatially organized in a grid \cite{kohonen2001}. 

\begin{figure}[htbp]
\includegraphics[width=\textwidth]{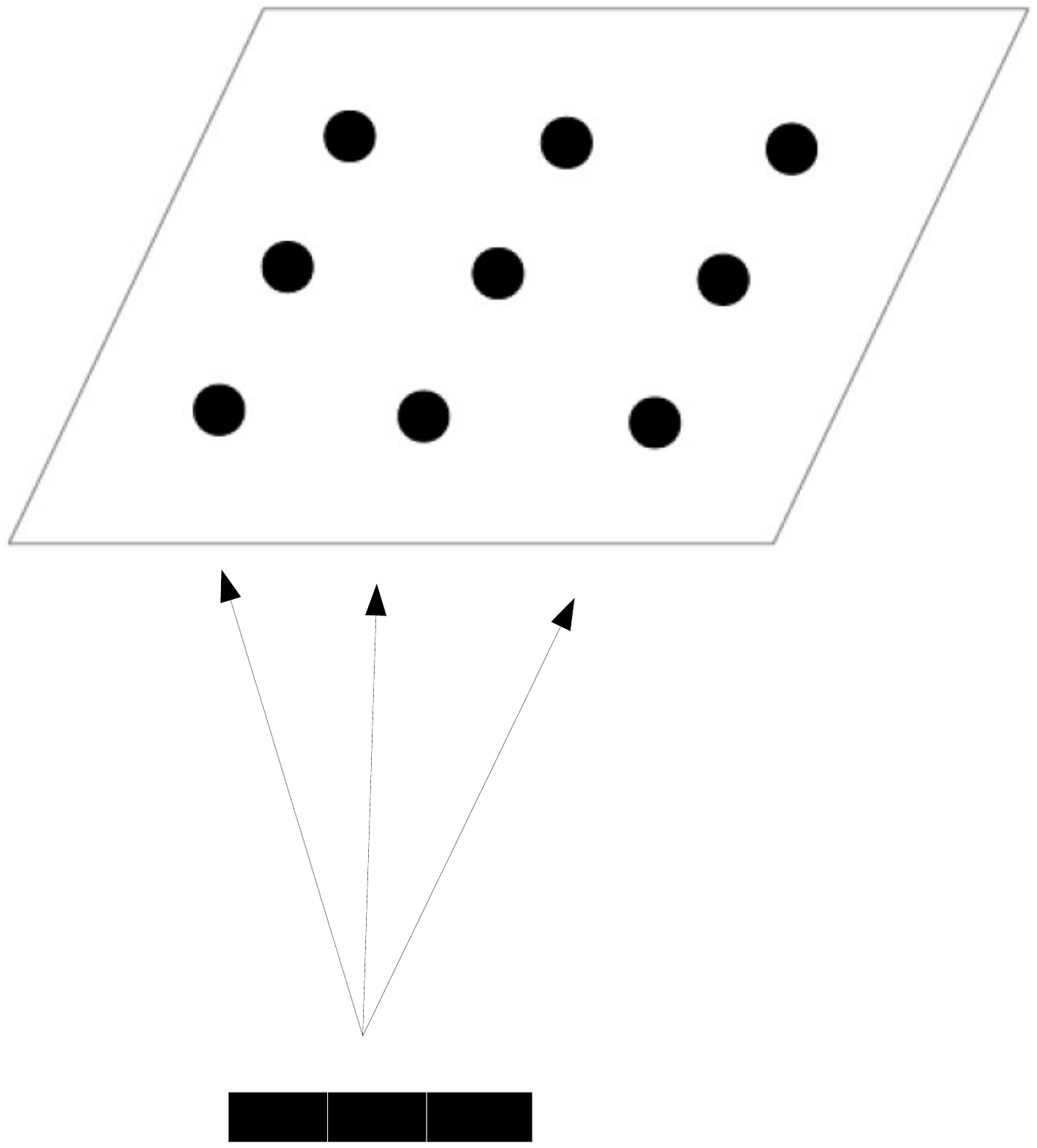}\caption{An example of SOM. The set of rectangles stands for the input presented to the SOM (in the example the input is three-dimensional). This is presented to {\em all} neurons of the SOM (these are the neurons-dots-in the upper grid) in order to find the $BMU$.}
\end{figure}

Each map unit $u$ is associated with a
weight vector $w_u$ of the same dimensionality as the input vectors.  
At the beginning of training, all weight vectors are initialized to random values, outside the range of values of the input stimuli. 
During training, the input elements  are sequentially presented to all neurons of the map. After each presentation of an input $x$, the {\em best-matching unit} (BMU$_x$) is selected: this is the unit $i$ whose weight vector $w_i$ is closest to the stimulus $x$ (i.e. $i = \arg\min_j\|x - w_j\|$).

The weights of the best matching unit and of its surrounding units are updated in order to maximize the chances that the same unit (or its surrounding units) will be selected as the best matching unit for the same stimulus or for similar stimuli {on subsequent presentations}. In particular,  it reduces the distance between the best matching unit's weights (and its surrounding neurons' weights) and the incoming input. Furthermore, it organizes the map topologically so that the weights of close-by neurons are updated in a similar direction, and come to react to similar inputs. We refer to \cite{kohonen2001} for the details.

%
%
%
%
%

The learning process is incremental: after the presentation of each input, the map's representation of the input (and in particular the representation of its best-matching unit) is updated in order to take into account the new incoming stimulus.
%
At the end of the whole process, the SOM has learned to organize the stimuli in a topologically significant way: similar inputs (with respect to Euclidean distance) are mapped to close by areas in the map, whereas inputs which are far apart from each other are mapped to distant areas of the map.

Once the SOM has learned to categorize, to assess category generalization, Gliozzi and Plunkett \cite{CogSci2017} define the map's disposition to consider a new stimulus $y$ as a member of a known category $C$ as a function of the {\em distance} of $y$ from the {\em map's representation} of $C$.
They take a minimalist notion of what is the map's category representation: this is the ensemble of best-matching units corresponding to the known instances of the category. 
They use $BMU_{C}$ to refer to the map's representation of category $C$ and define category generalization as depending on two elements:
\begin{itemize} 
\item the distance of the new stimulus $y$ with respect to the category representation 
\item {\em compared to} the maximal distance from that representation of all known instances of the category
\end{itemize} 
This captured by the following notion of {\em relative distance} ({\em rd} for short)  \cite{CogSci2017} :

\begin{equation}
\label{relative-distance}
rd(y,C) = \frac{min \|y - BMU_{C}\| }{max_{x \in C} \| x - BMU_{x}\| }
\end{equation}
where $min \|y - BMU_{C}\|$ is the (minimal) Euclidean distance between $y$ and $C$'s category representation, and ${max_{x \in C} \| x - BMU_{x}\| }$ expresses the {\em precision} of category representation,
and is the (maximal) Euclidean distance between any known member of the category and the category representation. 

With this definition, a given Euclidean distance from $y$ to $C's$ category representation will give rise to a higher {\em relative distance rd} if the maximal distance between C and its known examples is low (and category representation is precise) than if it is high (and category representation is coarse). 
As a function of the relative distance above, Gliozzi and Plunkett  then define the  {\em map's Generalization Degree} of category $C$ membership to a new stimulus $y$. 
%

It was observed that the above notion of relative distance (Equation \ref{relative-distance}) requires there to be a memory of some of the known instances of the category being used (this is needed to calculate the denominator in the equation). This gives rise to a sort of hybrid model in which category representation and some exemplars coexist. 
An alternative way of formulating the same notion of relative distance  would be to calculate {\em online} the distance between known category instance currently examined and the representation of the category being formed. 

By judging a new stimulus as belonging  to a category by comparing the distance of the stimulus from the category representation to the precision of the category representation, Gliozzi and Plunkett demonstrate  \cite{CogSci2017}  that the Numerosity and Variability effects of category generalization, described by Griffiths and Tenenbaum \cite{tengrif2001}, and usually explained with Bayesian tools, can be {accommodated} within a simple and psychologically plausible similarity-based account, which contrasts what was previously maintained.
In the next section, we show that their notion of relative distance can also be used as a basis for a logical semantics for SOMs.  

%


\section{Relating self-organising Maps and multi-preference models}

We aim at showing that, once the SOM has learned to categorize, we can regard the result of the categorization as a multipreference interpretation. 
Let $X$ be the set of input stimuli from different categories, $C_1, \ldots, C_k$, which have been considered during the learning process.

For each category $C_i$, we let $BMU_{C_i}$ be the ensemble of best-matching units corresponding to the input stimuli of category $C_i$, i.e.,
$BMU_{C_i}= \{ BMU_x \mid x \in X \mbox{ and } x \in C_i\}$.
We regard the learned categories $C_1, \ldots, C_k$ as being the concept names (atomic concepts) in the description logic and we let them constitute our§ set of distinguished concepts ${\cal C}= \{C_1, \ldots, C_k\}$. 

To construct a multi-preference interpretation we proceed as follows:
first, we fix the {\em domain} $\Delta^{s}$ to be the space of all possible stimuli; 
then,  for each category (concept) $C_i$,  we define a preference relation $<_{C_i}$, exploiting the notion of relative distance of a stimulus $y$ from the map's representation of $C_i$. Finally, we define the interpretation of concepts.

%
%
%
%
Let $\Delta^{s}$ be the set of all the possible stimuli, including all input stimuli ($X \subseteq \Delta^s$) as well as the best matching units of input stimuli (i.e., $\{BMU_x \mid x \in X \} \subseteq \Delta^s$).  For simplicity, we will assume that the space of input stimuli is finite.

Once the SOM has learned to categorize, the notion of relative distance $rd(x,C_i)$ of a stimulus $x$ from a category $C_i$ 
introduced above can be used to build 
a binary preference relation $<_{C_i}$ among the stimuli in $\Delta^s$ w.r.t. category $C_i$ as follows:  for all $x, x' \in \Delta^s$,
\begin{align}\label{preferenza_Ci}
x <_{C_i} x' \mbox{\ \  iff \ \ } rd(x,C_i) < rd(x' ,C_i)
\end{align}
Each preference relation $<_{C_i}$ is a strict partial order relation on $\Delta^s$.
The relation $<_{C_i}$ is also well-founded as we have assumed $\Delta^{s}$ to be finite. 

We exploit this notion of preference 
to define a concept-wise multipreference interpretation associated with the SOM, that we call a cw$^m$-model of the SOM.
We restrict the DL language to the fragment of $ \el$ (plus typicality) not admitting roles, 
as in the self-organising map we do not have a representation of role names.

 \begin{definition}[multipreference-model of a SOM]\label{modello_Som}  
The {\em multipreference-model of the SOM} is a multipreference interpretation 
$\emme^{s}= \langle \Delta^{s}, <_{C_1}, \ldots, <_{C_k}, \cdot^I \rangle$ 
such that:
\begin{itemize}
\item[(i)] $\Delta^{s}$ is the set of all the possible stimuli, as introduced above; 

\item[(ii)]
for each $C_i \in {\cal C}$, $<_{C_i}$ is the preference relation defined by equivalence (\ref{preferenza_Ci}).

\item[(iii)]  
 the interpretation function $\cdot^I$ is defined for concept names (i.e. categories) $C_i$ as follows:  
 $$C_i^I= \{y \in \Delta^s \mid rd(y,C_i) \leq rd_{max,C_i} \}$$
where $ rd_{max,C_i}$ 
is the maximal relative distance of an input stimulus $x \in C_i$ from category $C_i$, that is,
$rd_{max,C_i} = max_{x \in C_i} \{rd(x, C_i)\}$. 
The interpretation function $\cdot^I$ is extended to complex concepts 
according to Definition \ref{interpretazione_Ci}.

\end{itemize}
\end{definition}
Informally, we interpret  as $C_i$-elements those stimuli whose relative distance from category $C_i$ is not larger than the relative distance of any input exemplar belonging to category $C_i$.
Given $<_{C_i}$, we can identify the most typical $C_i$-elements  wrt $<_{C_I}$ 
as the $C_i$-elements whose relative distance from category $C_i$ is minimal, i.e., the elements in $min_{<_{C_i}}(C_i^I)$.
Observe that 
 the best matching unit $BMU_x$ of an input stimulus $x \in C_i$ is an element of $\Delta^s$.
Hence, for $y=BMU_x$,
the  relative distance of $y$ from category $C_i$, $rd(y,C_i)$, is $0$, as  $min \mid \mid y - BMU_{C_i} \mid \mid =0$.  
Therefore, 
$min_{<_{C_i}}(C_i^I) =\{ y \in \Delta^s \mid \; rd(y,C_i)=0\}$ and $BMU_{C_i} \subseteq min_{<_{C_i}}(C_i^I)$.

\subsection{Evaluation of concept inclusions by model checking} \label{sec:model_checking}

We have defined a multipreference interpretation $\emme^{s}$ where,  in the domain $\Delta^{s}$ of the possible stimuli, we are able to identify, for each category $C_i$,  the $C_i$-elements as well as the most typical $C_i$-elements wrt $<_{C_i}$.
We can exploit $\emme^s$ 
to verify which inclusions are satisfied by the SOM by {\em model checking}, i.e., by checking the satisfiability of inclusions over model $\emme^s$.  This can be done both for strict concept inclusions of the form $C_i \sqsubseteq C_j$ and for defeasible inclusions of the form $\tip(C_i) \sqsubseteq C_j$, where $C_i$ and $C_j$ are concept names (i.e., categories). 


For the verification that a typicality inclusion $\tip(C_i) \sqsubseteq C_j$ is satisfied in $\emme^s$ we have to check that the most typical $C_i$ elements wrt $<_{C_i}$ are $C_j$ elements, that is  $min_{<_{C_i}}(C_i^I) \subseteq C_j^I$. 
Note that, besides the elements in $BMU_{C_i} $, $min_{<_{C_i}}(C_i^I)$ may contain other elements of $\Delta^s$ having relative distance $0$ from $C_i$. 
As we do not know, for all the possible input stimuli in $\Delta^s$, whether they belong to $min_{<_{C_i}}(C_i^I)$ or to $ C_j^I$, 
as an approximation, we only check that all elements in  $BMU_{C_i} $ 
are $C_j$ elements, that is: 
\begin{equation} \label{cond_typ_incl}
\mbox{  for all  input stimuli $x \in C_i$, } rd(BMU_x, C_j) \leq rd_{max,C_j}
\end{equation}
Let the relative distance of $BMC_{C_i}$ from $C_j$ be defined as
$$rd( BMC_{C_i},C_j) = max_{x \in C_i} \{rd(BMU_x, C_j)\}$$
 the maximal relative distance of  $BMU_{C_i}$ from $C_j$. 
 Then we can rewrite condition (\ref{cond_typ_incl}) simply  as
$$rd( BMC_{C_i},C_j)  \leq  rd_{max,C_j}.$$
Observe that the relative distance $rd( BMC_{C_i},C_j)$ also gives a measure of plausibility of the defeasible inclusion $\tip(C_i) \sqsubseteq C_j$: the lower is the relative distance of $BMU_{C_i}$ from $C_j$, the more plausible is the defeasible inclusion $\tip(C_i) \sqsubseteq C_j$.

Verifying that a strict inclusion $C_i \sqsubseteq C_j$ is satisfied, requires to check that $C_i^I$ is included in $C_j^I$. Exploiting the fact that the map is organized topologically, and using the relative distance $rd( BMC_{C_i},C_j)$ of $BMC_{C_i}$ from $C_j$,
we verify that the relative distance of $BMC_{C_i}$ from $C_j$ plus the maximal relative distance of a $C_i$-element from  $C_i$ is not greater than the maximal relative distance of a  $C_j$-element from $C_j$:
\begin{equation} \label{cond_strict_incl}
rd( BMC_{C_i},C_j) + rd_{max,C_i} \leq rd_{max,C_j}
\end{equation}
where $rd_{max,C} = max_{y \in C} \{ rd(y, C)\}$.
That is, the $C_i$-element most distant  from $C_j$ is nearer to $C_j$ than the most distant $C_j$-element.

Computing conditions (\ref{cond_typ_incl}) and (\ref{cond_strict_incl}) on the SOM, may be non trivial, depending on the number of input stimuli that have been considered in the learning phase (the size of the set $X$ of input exemplars). However, from a logical point of view, this is just model checking.
Gliozzi and Plunkett have  considered self-organising maps that are able to learn from a limited number of input stimuli, although this is not generally true for all self-organising maps \cite{CogSci2017}.

\subsection{Combining preferences 
into a preferential interpretation}

The multipreference interpretation $\emme^s$ introduces in Definition \ref{modello_Som} allows to determine the set of $C_i$-elements for all learned categories $C_i$
and to define the most typical $C_i$-elements, exploiting the preference relation $<_{C_i}$. 
However, we are not able to define the most typical $C_i \sqcup C_j$-elements just using a single preference.
Starting from $\emme^s$, 
we construct a concept-wise multipreference interpretation $\emme^{som}$ that combines the preferential relations in $\emme^{s}$ into a global preference relation $<$, and provides an intepretation to  all  typicality concepts such as, for instance, $\tip(C_i \sqcap C_j \sqcap C_h)$.
The interpretation $\emme^{som}$ is constructed from $\emme^s$ according to Definition \ref{def-multipreference-int}.

The construction exploits a notion of specificity. Observe that the specificity relation between two concepts $C_i$ and $C_j$ can be determined based on the single model $\emme^s$ of the SOM. $C_i \succ C_j$  if $C_i \sqsubseteq C_j$ is satisfied in $\emme^s$ and $C_j \sqsubseteq C_i$ is not satisfied in $\emme^s$.

 \begin{definition}[cw$^m$-model of a SOM]\label{cwm_modello_Som}  
The {\em cw$^m$-model of a SOM} is a cw$^m$-interpretation 
$\emme^{som}= \langle \Delta^{s}, <_{C_1}, \ldots, <_{C_k}, <, \cdot^I \rangle$, 
such that the tuple $\langle \Delta^{s}, <_{C_1}, \ldots, <_{C_k}, \cdot^I \rangle$ is a multipreference model of the SOM 
according to Definition  \ref{modello_Som}, and $<$ is the global preference relation defined from $ <_{C_1}, \ldots, <_{C_k},$ according 
as in Definition \ref{def-multipreference-int}, point (c).
\end{definition}

In particular, in $\emme^{som}$, as in all cw$^m$-interpretations (see Definition  \ref{def-multipreference-int}), the interpretation of typicality concepts $\tip(C)$ is defined based on the global preference relation $<$ as $(\tip(C))^I= min_<(C^I)$, for all  concepts $C$. Here, we are considering concepts in the fragment of $\el$ language without roles, which are built from the concept names $C_1, \ldots, C_n$ (the learned categories).
The model $\emme^{som}$ can be considered a sort of (unique) canonical model for the SOM, representing what holds in that state of the SOM (e.g., after the learning phase).
The logical inclusions that ``follow from the SOM" are therefore the inclusions that hold in the single model $\emme^{som}$ (the situation is similar to the case of Horn clauses, where there is a unique minimal canonical model describing all the (atomic) logical consequences of the knowledge base).

As $\emme^{som}$ is a cw$^m$-interpretation, the result that the triple $ \langle \Delta^{s}, <, \cdot^I \rangle$ is a preferential interpretation as in KLM approach \cite{KrausLehmannMagidor:90,whatdoes} holds for $\emme^{som}$, and tells us that the model $\emme^{som}$ provides a logical semantics for the SOM which is  well-defined, as $\emme^{som}$ is a preferential consequence relation, and therefore satisfies all KLM properties of a preferential consequence relations. 

The verification of arbitrary defeasible inclusions on $\emme^{som}$  can, in principle, be done by model checking, but might require to consider all the possibly many input stimuli, i.e., all domain elements in $\Delta^s$, which may be unfeasible in practice.
As an alternative, 
the identification of the set of strict and defeasible inclusions satisfied by the SOM  over the learned categories $C_1, \ldots, C_k$ (as done in Section \ref{sec:model_checking}), allows to define an $\el$ 
knowledge base $K$ and to reason on it symbolically, using for instance 
an approach similar to the one described in Section \ref{sec:multipref} for ranked knowledge bases.
In particular, Answer Set Programming (in particular, {\em asprin}) has been used to achieve defeasible reasoning under the multipreference approach for the lightweight description logic $\elpb$  \cite{arXiv_iclp2020} . Ranked knowledge bases have been considered, where defeasible inclusions are given a rank, that provides a measure of plausibility of the defeasible inclusion, 
and multipreference entailment is reformulated as a problem of computing preferred answer sets. As we have seen, a measure of plausibility can as well be assigned to the defeasible inclusions satisfied by the SOM.


\subsection{Category generalization process as iterated belief revision}

We have seen that one can give an interpretation of a self-organising map after the learning phase, as a preferential model.
However, the state of the SOM during the learning phase can as well be represented as a multipreference model (precisely in the same way).
During training, the current state of the SOM corresponds to a model representing the beliefs about the input stimuli considered so far (beliefs concerning the category of the stimuli).

The category generalization process can then be regarded as a model building process and, in a way, as a belief revision process.
Initially we do not know  the category of the stimuli in the domain $\Delta^s$.
In the initial model, call it $\emme^{som}_0$ (over the domain $\Delta^s$) the interpretation of each concept $C_i$ is empty. 
$\emme^{som}_0$ is the model of a knowledge base $K_0$ containing a strict inclusion $C_i \sqsubseteq \bot$, for all $C_i$.

Each time a new input stimulus ($x \in C_i$) is considered, the model is revised adding the stimulus $x$ (and its best matching unit $BMU_{x}$) into the proper category ($C_i$).
Not only the category interpretation is revised by the addition of $x$ and $BMU_{x}$ in $C_i^I$ (so that $C_i \sqsubseteq \bot$ does not hold any more), but also the associated preference relation $<_{C_i}$ is revised as 
the addition of $BMU_{x}$ modifies the set of best matching units $BMU_{C_i}$ for category $C_i$, as well as the relative distance $rd(y,C_i)$ of a stimulus $y$ from $C_i$.  That is, a revision step may change the set of conditionals which are satisfied by the model.

If learning phase converges to a solution, the final state of the SOM is captured by the model $\emme^{som}$ obtained by a sequence of revision steps which, starting  from  $\emme^{som}_0$, gives rise to a sequence of models $\emme^{som}_0$,$\emme^{som}_{i_1}, \ldots$, $\emme^{som}_{i_r}$ (with $\emme^{som}=\emme^{som}_{i_r}$).
At each step  the knowledge base is not represented explicitly, but the model  $\emme^{som}_{i_j}$ of the knowledge base at step $j$ is used to determine the model at step $j+1$ as a result of revision ($\emme^{som}_{i_{j+1}}=\emme^{som}_{i_j} \star C_{i_j}(x_{i_j}) $).
The knowledge base $K$ (the set of all the strict and defeasible inclusions satisfied in $\emme^{som}$, can then be regarded as the knowledge base obtained from $K_0$ through a  sequence of revision steps, i.e., $K=K_0 \star C_{i1}(x_{i1}) \star \ldots \star C_{ir}(x_{ir})$.
In fact, from any state of the SOM we can construct a corresponding model, which determines a knowledge base, the set of (strict and defeasible) inclusions satisfied in that model. 
It would be interesting to study of the properties of  this notion of revision and compare with the notions of iterated belief revision studied in the literature \cite{DP97,GiordanoSL2002,Kern-IsbernerAMAI2004}. 

\section{Conclusions}
We have explored the relationships between a concept-wise multipreference semantics and self-organising maps. 
On the one hand, we have seen that self-organising maps can be given a logical semantics in terms of KLM-style preferential interpretations; 
the model can be used to learn or to validate conditional knowledge from the empirical data used in the category generalization process based on model checking; the learning process in the self-organising map can be regarded as an iterated belief revision process. 
On the other hand, the plausibility of concept-wise multipreference semantics is supported by that fact that self-organising maps 
are considered as psychologically and biologically plausible neural network models.

\end{document}